\documentclass[runningheads]{llncs}
\usepackage{graphicx}
\usepackage{upgreek}
\usepackage{amsmath}
\usepackage{amsfonts}
\usepackage{algorithm}
\usepackage{algpseudocode}
\usepackage{hyperref}
\usepackage{algorithmicx}
\usepackage[font=small,labelfont=bf]{caption}
\usepackage{color}
\usepackage[normalem]{ulem}
\usepackage{rotating}
\usepackage{tabularray}
\usepackage{booktabs}
\usepackage{graphicx}
\usepackage{setspace}
\usepackage{wrapfig}

\begin{document}
\title{Inductive Linear Probing for Few-shot Node Classification}
%
%
\author{Hirthik Mathavan\orcidID{0009-0003-6633-5866}
Zhen Tan\orcidID{0009-0006-9548-2330} \newline Nivedh Mudiam\orcidID{0009-0000-1133-1401} \and
Huan Liu\orcidID{0000-0002-3264-7904}}
\authorrunning{Hirthik Mathavan et al.}
%
\institute{Computer Science and Engineering, Arizona State University, Tempe, AZ 85287, USA\\
\email{\{hmathava,ztan36,nmudiam,huanliu\}@asu.edu}}
\maketitle              
\begin{abstract}
Meta-learning has emerged as a powerful training strategy for few-shot node classification, demonstrating its effectiveness in the transductive setting. However, the existing literature predominantly focuses on transductive few-shot node classification, neglecting the widely studied inductive setting in the broader few-shot learning community. This oversight limits our comprehensive understanding of the performance of meta-learning based methods on graph data. In this work, we conduct an empirical study to highlight the limitations of current frameworks in the inductive few-shot node classification setting. Additionally, we propose a simple yet competitive baseline approach specifically tailored for inductive few-shot node classification tasks. We hope our work can provide a new path forward to better understand how the meta-learning paradigm works in the graph domain.

\end{abstract}
\section{Introduction}



Graphs have found extensive applications across various research fields, including social network analysis\cite{14}, bioinformatics\cite{16}, recommendation systems\cite{13}, and more. Graphs are crucial in understanding user interactions, sentiment analysis, and community detection in social media mining. For example, consider a scenario where we aim to classify user's sentiments towards a particular product or event on a social media platform. The graph can represent users as nodes and their connections as edges, capturing their relationships and interactions. By analyzing the structural properties of the graph, such as user connections, and incorporating node attributes like past sentiments or textual content, node classification algorithms can assign sentiment labels to new, unlabeled users. However, getting labeled data for node classification can take time and effort in real-world scenarios. Few-shot learning, a sub-field of machine learning, attempts to address this issue by creating a model using just a few examples. Few-shot learning has gained significant interest lately because of its capability to learn swiftly from a restricted amount of labeled data. 

In recent years, meta-learning, also known as learning to learn, has emerged as a powerful technique for few-shot learning. Meta-learning involves training a model on a variety of tasks to learn a set of shared parameters that can be quickly adapted to new tasks with limited labeled data. In the context of graph node classification, meta-learning\cite{6} has been used to train models that can quickly adapt to new graphs with a few labeled examples.

While meta-learning has demonstrated promising results in the field of few-shot node classification\cite{19}, most of the existing works have focused on the transductive setting, where the graph neural network (GNN) encoder is trained and evaluated on the same graph. The inductive setting, where the model is trained on a set of graphs and tested on a new, unseen graph, has received less attention in the few-shot learning community. 
Also, due to the message passing mechanism, where nodes exchange information with their neighboring nodes to update their own representations in GNNs, the inductive setting poses additional challenges compared to the transductive setting. Consider the example of sentiment analysis described before. In an inductive setting, we encounter new social media platforms or events where we need to classify user sentiment without access to the entire graph used during training. This reflects the reality of dealing with evolving social media platforms and ever-changing user dynamics.

Inductive few-shot learning allows us to train a model on a diverse set of graphs and test its performance on unseen graphs, mimicking the real-world scenario where we encounter novel contexts. This emphasizes the importance of studying and developing effective few-shot learning approaches in the inductive setting, enabling models to adapt and make accurate predictions in dynamic real-world environments. Therefore, this work aims to bridge this gap by providing a comprehensive study of meta-learning for few-shot node classification in the inductive setting. We empirically show that most current meta-learning frameworks cannot perform well in this setting. Additionally, we introduce a straightforward yet effective baseline approach for inductive few-shot node classification tasks.


\section{Related Work}
In this section, we present an comprehensive review of the current literature concerning few-shot node classification and meta-learning, with a specific focus on the transductive setting.

\subsection{Few-shot Learning}

Few-shot learning (FSL) is a machine learning paradigm that serves to address concerns of limited data by capitalizing on knowledge gained from previous training data. Some example of models that employ FSL are Model-Agnostic Meta-Learning (MAML), Prototypical Networks, and Meta-GNN. 

MAML~\cite{3}  tackles the few-shot learning problem by learning an optimal initialization of model parameters. It enables fast adaptation to new tasks with limited examples through a two-step process: an inner loop for task-specific updates and an outer loop for optimizing adaptation across tasks. By iteratively fine-tuning the parameters, MAML achieves effective generalization and enables efficient few-shot learning across various domains. Prototypical Networks~\cite{2} capture the essence of similarities and dissimilarities among instances through a metric-based approach by computing class prototypes based on support examples and using distance-based classification. This approach enables accurate classification in few-shot scenarios which over various domains offers a valuable approach to few-shot learning tasks. Meta-GNN~\cite{4} instead primarily addresses few-shot learning when provided with graph structured data. The model enhances the capability of GNNs to capture expressive node representations and effectively generalize to new classes or tasks with limited labeled data.

\subsection{Meta Learning}
In the context of few-shot node classification, meta-learning algorithms have been proposed to learn effective representations and update strategies for handling new, unseen classes with only a few labeled examples. Popular meta-learning algorithms for few-shot learning include GPN, G-Meta etc.

Graph Prototypical Network (GPN)~\cite{6} introduces graph prototypes, learned through iterative aggregation with GNNs, as representative embeddings from the support set. By utilizing these prototypes, GPN achieves accurate few-shot classification by computing similarity scores between query nodes and prototypes. GPN's incorporation of graph-level information and iterative aggregation enables effective generalization and robust few-shot classification on graph-structured data. G-Meta~\cite{5} combines subgraph extraction with GNNs to learn expressive node representations. It employs the MAML strategy to iteratively update and meta-update GNN parameters. This enables efficient adaptation to new tasks and improved classification on query nodes. Other models like AMM-GNN extend MAML with an attribute matching mechanism, and TENT reduces the variance among different meta-tasks for better generalization performance. Existing works primarily focus on transductive few-shot node classification, neglecting the widely studied inductive setting. We empirically evaluate meta-learning frameworks in the inductive setting to gain deeper insights into their performance on graphs.

\section{Preliminaries}
\subsection{Problem Statement}
The problem of few-shot node classification is concerned with attributed networks represented as $G = (\mathcal{V}, \mathcal{E},X) = (A, X)$, where $V$ is the set of nodes $v_1, v_2, \ldots, v_n$, $\mathcal{E}$ is the set of edges $e_1, e_2, \ldots, e_m$ , $X = [x_1; x_2; \ldots; x_n] \in \mathbb{R}^{n \times d}$ is the matrix of node features, and $A = \{0, 1\}^{n \times n}$ is the adjacency matrix representing the network structure. Each element in $A$ is either 0 or 1, indicating the absence or presence of an edge between nodes. The task involves a series of node classification tasks $T = {\{T_i\}}_{i=1}^I$, where $T_i$ is a dataset for a particular task, and $I$ is the number of such tasks. The classes of nodes available during training are referred to as base classes, while the classes during the target test phase are referred to as novel classes, and the intersection of the two sets is empty. Notably, under different settings, labels of nodes for training (i.e., $C_{base}$) may or may not be available during training. Conventionally, there are few labeled nodes for novel classes $C_{novel}$ during the test phase.  

\textbf{Definition 1. Few-shot Node Classification (FSNC):} Few-shot node classification refers to a problem in which an attributed graph $G = (A,X)$ is given, with a label space C divided into two sets, $C_{base}$ and $C_{novel}$. The goal is to predict the labels of unlabeled nodes (query set Q) from $C_{novel}$, given only a few labeled nodes (support set S) for $C_{novel}$. If each task in the test set has N novel classes and K labeled nodes for each class, then this task is referred to as an N-way K-shot node classification problem. 

\begin{wrapfigure}{i}{0.5\textwidth}
\vspace{-20pt}
    \centering
    \includegraphics[scale=0.4]{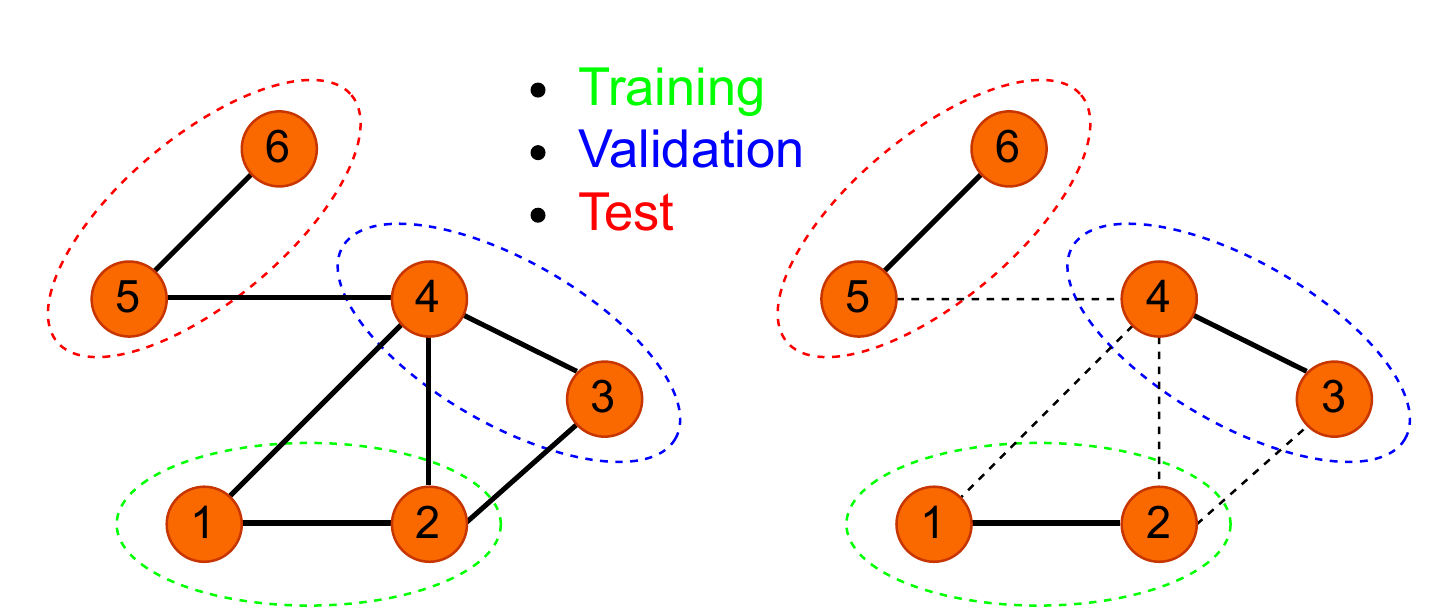}
    \caption{Transductive/Inductive Setting}
    \label{trans}
    \vspace{-20pt}
\end{wrapfigure}

\textbf{Transductive Setting:} In the transductive setting, the input graph is observed in all dataset splits, including the training, validation, and test sets (Fig.~\ref{trans}). The graph remains intact, and only the node  labels are split for training and evaluation purposes.  During training, embeddings are computed using the entire graph, and the model is trained using the labels of selected nodes (e.g., node 1 and node 2). During validation, embeddings are again computed using the entire graph, and the model's performance is evaluated on the labels of other nodes (e.g., node 3 and node 4).

\textbf{Inductive Setting:} In the inductive setting, the graph is modified by breaking the edges between the dataset splits, resulting in different neighbor environments for nodes compared to the transductive setting (Fig.~\ref{trans}). For example, node 4 will no longer have an influence on the prediction of node 1. During training, embeddings are computed using the graph specific to the training split, such as the graph over node 1 and node 2. The model is trained using the labels of these selected nodes. During validation, embeddings are computed using the graph specific to the validation split, such as the graph over node 3 and node 4. The model's performance is then evaluated on the labels of these respective nodes (node 3 and node 4). This will further lead to the change of message passing, making it harder for GNNs to learn generalizable knowledge~\cite{16}. 
\subsection{Episodic Meta-learning for FSNC}
Episodic meta-learning has emerged as an effective paradigm for addressing few-shot learning tasks, garnering substantial attention~\cite{21,22}.
The underlying concept of episodic meta-learning involves training neural networks to mimic the evaluation conditions, which is believed to improve prediction performance on test tasks~\cite{21,22}. This paradigm has been successfully extended to few-shot node classification in the graph domain, as demonstrated by recent works~\cite{6,19,25}.
In the context of few-shot node classification, the training phase follows a specific procedure. Meta-train tasks or episodes, denoted as $T_{tr}$, are generated from a base class set $C_{base}$, to emulate the test tasks. These episodes adhere to N-way K-shot node classification specifications. Each episode, denoted as $T_t$, comprises a support set $S_t$, and a query set $Q_t$, defined as follows:
\begin{equation}
\begin{aligned}
&T_{tr} = \{T_t\}_{t=1}^\mathcal{T} = \{ T_1, T_2,...,T_\mathcal{T}\}, \\
&T_t = \{S_t, Q_t\},\\
&S_t = \{(v_1, y_1), (v_2, y_2), \ldots, (v_{N \times K}, y_{N \times K})\}, \\
&Q_t = \{(v_1, y_1), (v_2, y_2), \ldots, (v_{N \times K}, y_{N \times K})\}. \\
\end{aligned}
\end{equation}

In a typical meta-learning method, within each episode, K labeled nodes are randomly sampled from N base classes to form the support set. This support set is then used to train a GNN model, simulating the N-way K-shot node classification scenario during the test phase. Subsequently, the GNN predicts labels for a query set, which comprises nodes randomly sampled from the same classes as the support set. The optimization process involves minimizing the Cross-Entropy Loss ($L_{CE}$) w.r.t. the GNN encoder $g_\theta$ and the classifier $f_\phi$:
\begin{equation}
\begin{aligned}
&\theta, \psi = \operatorname*{arg\ min}_{\theta,\psi} L_{CE}(T_{t}; \theta, \psi).
\end{aligned}
\end{equation}

Several approaches have been proposed based on this framework such as Meta-GNN\cite{4}, GPN\cite{6}, G-Meta\cite{5} etc. Nevertheless, the evaluation of these methods has predominantly been conducted under transductive settings, neglecting the exploration of their performance in inductive settings. 
\subsection{Proposed Baseline}
Our work is motivated by the Intransigent GNN model (I-GNN) introduced by a previous study\cite{20}. The I-GNN model proposes a straightforward approach for few-shot learning that relies on reusing features instead of using complex meta-learning algorithms to achieve fast adaptation. The authors show that the I-GNN model, despite its simplicity, can achieve competitive performance compared to meta-learning based approaches. In our study, we adapt the I-GNN model to the inductive setting and propose a simple yet effective baseline for inductive few-shot node classification tasks.

The I-GNN model is designed to be inflexible and unadaptable to new tasks. The training process of I-GNN is split into two phases. In the first phase, a GNN encoder ($g_\theta$) and a linear classifier ($f_\phi$) are pre-trained on all base classes ($C_{base}$) using vanilla supervision through the $L_{CE}$ loss function. A weight-decay regularization term is also applied during this phase. In the second phase, the parameter of the GNN encoder is frozen, and the classifier is discarded. When fine-tuning on a target few-shot node classification task, the pretrained GNN encoder is used to directly transfer embeddings of all nodes from the task, and a new linear classifier ($f_\psi$) is involved and tuned with few-shot labeled nodes from the support set ($S_i$) to predict labels of nodes in the query set ($Q_i$).
\begin{equation}
\begin{aligned}
&T_{tr}^{'} = \cup\{T_t\}_{t=1}^\mathcal{T} = \cup\{ T_1, T_2,...,T_\mathcal{T}\} \\
&\theta, \phi = \operatorname*{arg\ min}_{\theta,\phi} L_{CE}(T_{tr}^{'}; \theta, \phi) + R(\theta),
\end{aligned}
\end{equation}
\begin{equation}
\psi = \operatorname*{arg\ min}_\psi L_{CE}(S_i;\theta,\psi)
\end{equation}
\section{Empirical Evaluation}
\subsection{Experimental Settings}
In this research study, various methods for few-shot node classification are evaluated through systematic experiments under the inductive setting. These methods include ProtoNet\cite{2}, MAML\cite{3}, Meta-GNN\cite{4}, G-Meta\cite{5}, GPN\cite{6}, AMM-GNN\cite{7}, and TENT\cite{8}. The performance of these methods is compared on five real-world graph datasets: CoraFull\cite{9}, Coauthor-CS\cite{11}, Amazon-Computer\cite{11}, Cora\cite{12}, and CiteSeer\cite{12}.
\begingroup
\vspace{-25pt}
\renewcommand{\arraystretch}{1} 
\begin{table}[!htb]
\caption{Statistics of Benchmark Datasets}\label{tab1}
\centering
\begin{tabular}{cccccccc}
\hline
\textbf{Dataset} & \# Nodes & \# Edges & \# Features & $|C|$ & $|C_{train}|$ & $|C_{dev}|$ & $|C_{test}|$
\\ \hline
CoraFull        & 19,793  & 63,421    & 8,710 & 70 & 40 & 15 & 15 
\\
Coauthor-CS     & 18,333  & 81,894    & 6,805 & 15 & 5  & 5  & 5  \\
Amazon-Computer & 13,752  & 245,861   & 767   & 10 & 4  & 3  & 3  \\
Cora            & 2,708   & 5,278     & 1,433 & 7  & 3  & 2  & 2  \\
CiteSeer        & 3,327   & 4,552     & 3,703 & 6  & 2  & 2  & 2  \\ \hline

\end{tabular}
\vspace{-15pt}
\end{table}
\endgroup

CoraFull, Coauthor-CS, Amazon-Computer, Cora, and CiteSeer are five prevalent real-world graph datasets, each consisting of multiple node classes for training and evaluation. These datasets include citation networks, co-authorship graphs, and co-purchase graphs, and the task is to predict the category of a certain publication or paper. The number of node classes used for training, development, and testing varies depending on the dataset.  Table~\ref{tab1} describes the statistics of the datasets.

\subsection{Evaluation Protocol}
This section outlines the evaluation protocol used to compare the meta-learning methods. The node label space $C$ of an graph dataset $G = (A,X)$ is divided into $\{C_{base}, C_{novel} \text{ or } C_{test}\}$. $C_{base}$ is split into $C_{train}$ and $C_{dev}$ (division strategy for each dataset are in Table ~\ref{tab1}). Evaluation is done by providing a GNN encoder $g$, a classifier, $f$, an epoch interval $EI$ for validation, $S$ sampled meta-tasks for evaluation, $E$ epoch patience, $M$ maximum epoch number, $T$ experiment repeated times, and $N$-way $K$-shot, $Q$-query settings specification. The algorithm~\ref{algo1} calculates the final FSNC accuracy $\mathcal{A}$ and confident interval $\mathcal{CI}$. The default values of all the parameters are as follows, $EI = 10; S=100; E=10; M=10000; T=5; N=\{2,5\}; K=\{1,3,5\}; Q=10$.
\vspace{-15pt}

\algrenewcommand\alglinenumber[1]{\tiny #1:}
\begin{algorithm}[!htb]
\captionsetup{font=scriptsize}
\caption{UNIFIED EVALUATION PROTOCOL FOR FEW-SHOT NODE CLASSIFICATION}\label{alg:cap}
\tiny
\begin{algorithmic}[1]
\renewcommand{\algorithmicrequire}{\textbf{Input:}}
\Require {$\text{Graph } G,\text{ } C_{train}, C_{dev}, C_{test}; \text{ GNN } g, \text{ classifier } f; \text{ parameters } EI = 10 , S=100, E=10, M=10000, T=5, N=2,5, K=1,3,5, Q=10$} 
\renewcommand{\algorithmicensure}{\textbf{Output:}}
\Ensure $f, \text{accuracy } \mathcal{A}, \text{confident interval } I, \text{trained models } g$ \newline
Repeat experiment for T times
\For{$i = 1, 2, . . . , T$}
\State $j \leftarrow 1, k \leftarrow 1, a_{best} \leftarrow 0$;
\While{$k \leq M$}
\State Optimize $g$ based on the specific training strategy; \Comment{Training}
\If{$k\ mod\ EI = 0$}{
\State Sample $S$ meta-tasks from $C_{dev}$ on $G$; \Comment{Validation}
\State Calculate the obtained few-shot node classification accuracy $a$;
\If{$a > a_{best}$}{ $a_{best} \leftarrow a, j \leftarrow 0$;
}
\Else{
$j \leftarrow j + 1$;
}
\EndIf
}
\EndIf
\If{$j = E$}{
break; \Comment{Early Break}
}
\EndIf
\EndWhile
\State Sample $S$ meta-tasks from $C_{test}$ on $G$; \Comment{Test}
\State Calculate the obtained classification accuracy $a_{test}$;
\State $a_{r} \leftarrow a_{test}, i \leftarrow i + 1$;
\EndFor
\State Calculate averaged accuracy $\mathcal{A}$ and confident interval $\mathcal{CI}$ based on $\{{a_{1}, a_{2}, . . . , a_{i}}\}$;
\end{algorithmic}
\label{algo1}
\end{algorithm}

\subsection{Comparison}
\begin{figure}[H]
    \centering
    \includegraphics[width=12 cm]{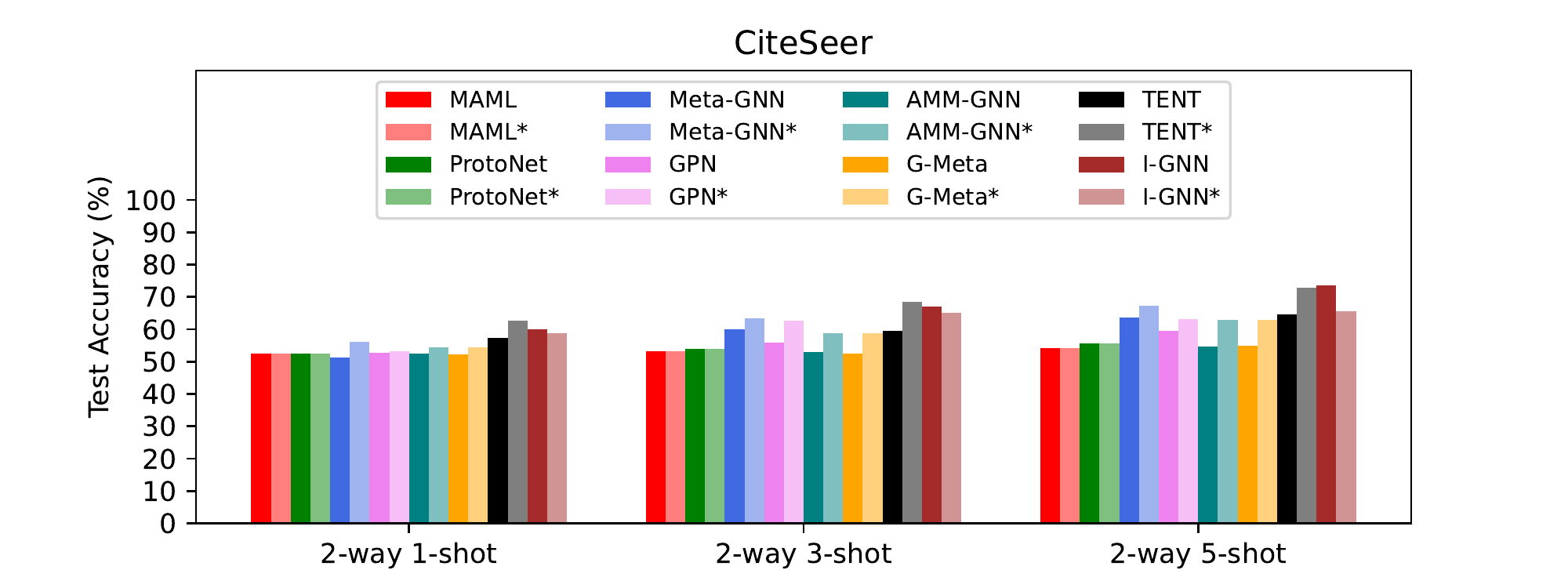}
    \caption{Meta-Learning, I-GNN with inductive and transductive (*)}
    \label{fig1}
\end{figure}

In Table ~\ref{tab2}, the performance of different meta-learning methods and the proposed baseline is compared for few-shot node classification tasks. The comparison includes four distinct few-shot settings: 5-way 1-shot, 5-way 5-shot, 2-way 1-shot, and 2-way 5-shot, allowing for a comprehensive analysis. The evaluation metrics used are the average classification accuracy and the 95\% confidence interval, which are computed based on multiple repetitions (T). 
Fig ~\ref{fig1} presents the performance results of the CiteSeer dataset (similar trends observed in other datasets) for various N-way K-shot settings. The observations derived from the results are as follows:
\begin{table}[!htb]
\vspace{-20pt}
\caption{Few-shot node classification results of meta-learning methods and I-GNN. Accuracy ($\uparrow$) and Confidence Interval ($\downarrow$) are in \%. The best and second best results are bold and underlined, respectively.}\label{tab2}
\centering
\resizebox{\columnwidth}{!}{%
\begin{tabular}{@{}ccccccccccc@{}}
\toprule
\textbf{Dataset} &
  \multicolumn{2}{c}{\textbf{CoraFull}} &
  \multicolumn{2}{c}{\textbf{Coauthor-CS}} &
  \multicolumn{2}{c}{\textbf{Cora}}   &
  \multicolumn{2}{c}{\textbf{Amazon-Computer}} &
  \multicolumn{2}{c}{\textbf{CiteSeer}} \\ \midrule
\multicolumn{1}{|c|}{\textbf{Settings}} &
  \multicolumn{1}{c|}{\textbf{5-way 1-shot}} &
  \multicolumn{1}{c|}{\textbf{5-way 5-shot}} &
  \multicolumn{1}{c|}{\textbf{5-way 1-shot}} &
  \multicolumn{1}{c|}{\textbf{5-way 5-shot}} &
  \multicolumn{1}{c|}{\textbf{2-way 1-shot}} &
  \multicolumn{1}{c|}{\textbf{2-way 5-shot}} &
  \multicolumn{1}{c|}{\textbf{2-way 1-shot}} &
  \multicolumn{1}{c|}{\textbf{2-way 5-shot}} &
  \multicolumn{1}{c|}{\textbf{2-way 1-shot}} &
  \multicolumn{1}{c|}{\textbf{2-way 5-shot}} \\ \midrule
\multicolumn{11}{c}{\textbf{Inductive}} \\ \midrule
\multicolumn{1}{|c|}{\textbf{MAML}} &
  \multicolumn{1}{c|}{22.63 $\pm$ 1.19} &
  \multicolumn{1}{c|}{27.21 $\pm$ 1.32} &
  \multicolumn{1}{c|}{27.98 $\pm$ 1.42} &
  \multicolumn{1}{c|}{42.12 $\pm$ 1.40} &
  \multicolumn{1}{c|}{53.13 $\pm$ 2.26} &
  \multicolumn{1}{c|}{57.39 $\pm$ 2.23} &
  \multicolumn{1}{c|}{52.67$\pm$2.11} &
  \multicolumn{1}{c|}{58.23$\pm$2.53} &
  \multicolumn{1}{c|}{52.39$\pm$2.20} &
  \multicolumn{1}{c|}{54.13$\pm$2.18} \\ \midrule
\multicolumn{1}{|c|}{\textbf{ProtoNet}} &
  \multicolumn{1}{c|}{32.43 $\pm$ 1.61} &
  \multicolumn{1}{c|}{51.54 $\pm$ 1.68} &
  \multicolumn{1}{c|}{32.13 $\pm$ 1.52} &
  \multicolumn{1}{c|}{49.25 $\pm$ 1.50} &
  \multicolumn{1}{c|}{53.04 $\pm$ 2.36} &
  \multicolumn{1}{c|}{57.92 $\pm$ 2.34} &
  \multicolumn{1}{c|}{\underline{61.98$\pm$2.95}} &
  \multicolumn{1}{c|}{\underline{70.20$\pm$2.64}} &
  \multicolumn{1}{c|}{52.51$\pm$2.44} &
  \multicolumn{1}{c|}{55.69$\pm$2.27} \\ \midrule
\multicolumn{1}{|c|}{\textbf{Meta-GNN}} &
  \multicolumn{1}{c|}{34.97 $\pm$ 1.78} &
  \multicolumn{1}{c|}{49.32 $\pm$ 1.99} &
  \multicolumn{1}{c|}{37.78 $\pm$ 2.02} &
  \multicolumn{1}{c|}{51.17 $\pm$ 1.91} &
  \multicolumn{1}{c|}{52.09 $\pm$ 2.39} &
  \multicolumn{1}{c|}{58.21 $\pm$ 1.52} &
  \multicolumn{1}{c|}{55.47$\pm$2.43} &
  \multicolumn{1}{c|}{59.12$\pm$2.55} &
  \multicolumn{1}{c|}{51.18$\pm$2.04} &
  \multicolumn{1}{c|}{63.68$\pm$2.65} \\ \midrule
\multicolumn{1}{|c|}{\textbf{GPN}} &
  \multicolumn{1}{c|}{27.90 $\pm$ 1.35} &
  \multicolumn{1}{c|}{36.40 $\pm$ 1.82} &
  \multicolumn{1}{c|}{35.00 $\pm$ 1.55} &
  \multicolumn{1}{c|}{49.30 $\pm$ 2.63} &
  \multicolumn{1}{c|}{50.00 $\pm$ 1.89} &
  \multicolumn{1}{c|}{55.00 $\pm$ 1.81} &
  \multicolumn{1}{c|}{49.75$\pm$0.85} &
  \multicolumn{1}{c|}{54.25$\pm$2.45} &
  \multicolumn{1}{c|}{52.75$\pm$1.85} &
  \multicolumn{1}{c|}{59.50$\pm$2.10} \\ \midrule
\multicolumn{1}{|c|}{\textbf{AMM-GNN}} &
  \multicolumn{1}{c|}{36.45 $\pm$ 1.99} &
  \multicolumn{1}{c|}{52.09 $\pm$ 1.90} &
  \multicolumn{1}{c|}{\underline{53.30 $\pm$ 2.39}} &
  \multicolumn{1}{c|}{\textbf{72.64 $\pm$ 1.48}} &
  \multicolumn{1}{c|}{\underline{54.36 $\pm$ 2.20}} &
  \multicolumn{1}{c|}{\underline{60.01 $\pm$ 2.40}} &
  \multicolumn{1}{c|}{51.99$\pm$1.51} &
  \multicolumn{1}{c|}{52.48$\pm$1.57} &
  \multicolumn{1}{c|}{52.40$\pm$2.14} &
  \multicolumn{1}{c|}{54.63$\pm$2.24} \\ \midrule
\multicolumn{1}{|c|}{\textbf{G-Meta}} &
  \multicolumn{1}{c|}{\underline{40.76 $\pm$ 2.19}} &
  \multicolumn{1}{c|}{\underline{57.69 $\pm$ 1.93}} &
  \multicolumn{1}{c|}{46.79 $\pm$ 1.95} &
  \multicolumn{1}{c|}{66.95 $\pm$ 1.43} &
  \multicolumn{1}{c|}{53.78 $\pm$ 2.05} &
  \multicolumn{1}{c|}{58.35 $\pm$ 2.15} &
  \multicolumn{1}{c|}{52.27$\pm$1.98} &
  \multicolumn{1}{c|}{61.03$\pm$2.19} &
  \multicolumn{1}{c|}{52.21$\pm$2.17} &
  \multicolumn{1}{c|}{54.92$\pm$2.26} \\ \midrule
\multicolumn{1}{|c|}{\textbf{TENT}} &
  \multicolumn{1}{c|}{38.90 $\pm$ 2.20} &
  \multicolumn{1}{c|}{54.32 $\pm$ 1.65} &
  \multicolumn{1}{c|}{\textbf{53.52 $\pm$ 1.73}} &
  \multicolumn{1}{c|}{\underline{68.16 $\pm$ 1.18}} &
  \multicolumn{1}{c|}{50.40 $\pm$ 2.01} &
  \multicolumn{1}{c|}{59.80 $\pm$ 2.38}  &
  \multicolumn{1}{c|}{\textbf{82.40$\pm$2.28}} &
  \multicolumn{1}{c|}{\textbf{92.00$\pm$1.18}} &
  \multicolumn{1}{c|}{\underline{57.35$\pm$2.74}} &
  \multicolumn{1}{c|}{\underline{64.55$\pm$2.63}} \\ \midrule
\multicolumn{1}{|c|}{\textbf{I-GNN}} &
  \multicolumn{1}{c|}{\textbf{47.14 $\pm$ 2.08}} &
  \multicolumn{1}{c|}{\textbf{59.01 $\pm$ 1.82}} &
  \multicolumn{1}{c|}{37.23 $\pm$ 1.70} &
  \multicolumn{1}{c|}{51.24 $\pm$ 1.42} &
  \multicolumn{1}{c|}{\textbf{62.33 $\pm$ 2.67}} &
  \multicolumn{1}{c|}{\textbf{70.16 $\pm$ 2.05}} &
   \multicolumn{1}{c|}{59.08$\pm$2.67} &
  \multicolumn{1}{c|}{68.35$\pm$2.48} &
  \multicolumn{1}{c|}{\textbf{60.04$\pm$1.55}} &
  \multicolumn{1}{c|}{\textbf{73.63$\pm$2.03}} \\ \midrule
\multicolumn{11}{c}{\textbf{Transductive}} \\ \midrule
\multicolumn{1}{|c|}{\textbf{MAML}} &
  \multicolumn{1}{c|}{22.63 $\pm$ 1.19} &
  \multicolumn{1}{c|}{27.21 $\pm$ 1.32} &
  \multicolumn{1}{c|}{27.98 $\pm$ 1.42} &
  \multicolumn{1}{c|}{42.12 $\pm$ 1.40} &
  \multicolumn{1}{c|}{53.13 $\pm$ 2.26} &
  \multicolumn{1}{c|}{57.39 $\pm$ 2.23} &
   \multicolumn{1}{c|}{52.67$\pm$2.11} &
  \multicolumn{1}{c|}{58.23$\pm$2.53} &
  \multicolumn{1}{c|}{52.39$\pm$2.20} &
  \multicolumn{1}{c|}{54.13$\pm$2.18} \\ \midrule
\multicolumn{1}{|c|}{\textbf{ProtoNet}} &
  \multicolumn{1}{c|}{32.43 $\pm$ 1.61} &
  \multicolumn{1}{c|}{51.54 $\pm$ 1.68} &
  \multicolumn{1}{c|}{32.13 $\pm$ 1.52} &
  \multicolumn{1}{c|}{49.25 $\pm$ 1.50} &
  \multicolumn{1}{c|}{53.04 $\pm$ 2.36} &
  \multicolumn{1}{c|}{57.92 $\pm$ 2.34} &
  \multicolumn{1}{c|}{61.98$\pm$2.95} &
  \multicolumn{1}{c|}{70.20$\pm$2.64} &
  \multicolumn{1}{c|}{52.51$\pm$2.44} &
  \multicolumn{1}{c|}{55.69$\pm$2.27} \\ \midrule
\multicolumn{1}{|c|}{\textbf{Meta-GNN}} &
  \multicolumn{1}{c|}{55.33 $\pm$ 2.43} &
  \multicolumn{1}{c|}{70.50 $\pm$ 2.02} &
  \multicolumn{1}{c|}{52.86 $\pm$ 2.14} &
  \multicolumn{1}{c|}{68.59 $\pm$ 1.49} &
  \multicolumn{1}{c|}{\underline{65.27 $\pm$ 2.93}} &
  \multicolumn{1}{c|}{72.51 $\pm$ 1.91} &
  \multicolumn{1}{c|}{65.19$\pm$3.29} &
  \multicolumn{1}{c|}{78.65$\pm$3.12} &
  \multicolumn{1}{c|}{56.14$\pm$2.62} &
  \multicolumn{1}{c|}{\underline{67.34$\pm$2.10}} \\ \midrule
\multicolumn{1}{|c|}{\textbf{GPN}} &
  \multicolumn{1}{c|}{52.75 $\pm$ 2.32} &
  \multicolumn{1}{c|}{72.82 $\pm$ 1.88} &
  \multicolumn{1}{c|}{60.66 $\pm$ 2.07} &
  \multicolumn{1}{c|}{\textbf{81.79 $\pm$ 1.18}} &
  \multicolumn{1}{c|}{62.61 $\pm$ 2.71} &
  \multicolumn{1}{c|}{76.39 $\pm$ 2.33} &
  \multicolumn{1}{c|}{57.26$\pm$1.50} &
  \multicolumn{1}{c|}{77.63$\pm$2.91} &
  \multicolumn{1}{c|}{53.10$\pm$2.39} &
  \multicolumn{1}{c|}{63.09$\pm$2.50} \\ \midrule
\multicolumn{1}{|c|}{\textbf{AMM-GNN}} &
  \multicolumn{1}{c|}{\underline{58.77 $\pm$ 2.49}} &
  \multicolumn{1}{c|}{\underline{75.61 $\pm$ 1.78}} &
  \multicolumn{1}{c|}{\underline{62.04 $\pm$ 2.26}} &
  \multicolumn{1}{c|}{\underline{81.78 $\pm$ 1.24}} &
  \multicolumn{1}{c|}{65.23 $\pm$ 2.67} &
  \multicolumn{1}{c|}{\textbf{82.30 $\pm$ 2.07}} &
  \multicolumn{1}{c|}{\underline{71.04$\pm$3.56}} &
  \multicolumn{1}{c|}{\underline{79.21$\pm$3.38}} &
  \multicolumn{1}{c|}{54.53$\pm$2.51} &
  \multicolumn{1}{c|}{62.93$\pm$2.42} \\ \midrule
\multicolumn{1}{|c|}{\textbf{G-Meta}} &
  \multicolumn{1}{c|}{\textbf{60.44 $\pm$ 2.48}} &
  \multicolumn{1}{c|}{\textbf{75.84 $\pm$ 1.70}} &
  \multicolumn{1}{c|}{59.68 $\pm$ 2.16} &
  \multicolumn{1}{c|}{74.18 $\pm$ 1.29} &
  \multicolumn{1}{c|}{\textbf{67.03 $\pm$ 3.22}} &
  \multicolumn{1}{c|}{\underline{80.05 $\pm$ 1.98}} &
  \multicolumn{1}{c|}{63.68$\pm$3.05} &
  \multicolumn{1}{c|}{70.21$\pm$3.16} &
  \multicolumn{1}{c|}{55.15$\pm$2.68} &
  \multicolumn{1}{c|}{64.53$\pm$2.35} \\ \midrule
\multicolumn{1}{|c|}{\textbf{TENT}} &
  \multicolumn{1}{c|}{55.44 $\pm$ 2.08} &
  \multicolumn{1}{c|}{70.10 $\pm$ 1.73} &
  \multicolumn{1}{c|}{\textbf{63.70 $\pm$ 1.88}} &
  \multicolumn{1}{c|}{76.90 $\pm$ 1.19} &
  \multicolumn{1}{c|}{53.05 $\pm$ 2.78} &
  \multicolumn{1}{c|}{62.15 $\pm$ 2.13} &
  \multicolumn{1}{c|}{\textbf{71.15$\pm$3.11}} &
  \multicolumn{1}{c|}{\textbf{79.25$\pm$2.61}} &
  \multicolumn{1}{c|}{\textbf{62.75$\pm$3.23}} &
  \multicolumn{1}{c|}{\textbf{72.95$\pm$2.13}} \\ \midrule
\multicolumn{1}{|c|}{\textbf{I-GNN}} &
  \multicolumn{1}{c|}{42.70 $\pm$ 1.92} &
  \multicolumn{1}{c|}{51.46 $\pm$ 1.69} &
  \multicolumn{1}{c|}{43.89 $\pm$ 1.82} &
  \multicolumn{1}{c|}{55.93 $\pm$ 1.46} &
  \multicolumn{1}{c|}{54.45 $\pm$ 3.13} &
  \multicolumn{1}{c|}{65.18 $\pm$ 2.21} &
  \multicolumn{1}{c|}{62.3$\pm$22.89} &
  \multicolumn{1}{c|}{72.81$\pm$2.93} &
  \multicolumn{1}{c|}{\underline{58.70$\pm$3.17}} &
  \multicolumn{1}{c|}{65.60$\pm$2.59} \\ \bottomrule
\end{tabular}%
}
\vspace{-20pt}
\end{table}
\begin{itemize}
  \item In the inductive setting, except for MAML and ProtoNet, meta-learning models exhibit a \textbf{significant performance drop} compared to the transductive setting. This decline is attributed to the challenges of generalizing knowledge from limited labeled examples to unseen data. In the transductive setting, models access the entire graph for predictions, while in the inductive setting, they must generalize to new nodes or graphs. Limited labeled data and the need for generalization contribute to lower performance in the inductive setting.
  \item I-GNN shows \textbf{superior performance} in the inductive setting compared to the transductive setting for certain datasets like Cora, Citeseer, and CoraFull. This can be due to its ability to capture more transferable node embedding in the inductive setting.
  \item  The scores for both MAML and ProtoNet \textbf{remain the same} on all datasets because they do not utilize message-passing GNN in their approach. Since they do not leverage the graph structure and operate on a per-node basis, the performance drop observed in other meta-learning models under the inductive setting does not affect them in the same way. 
  Therefore, their performance remains consistent between the transductive \& inductive settings. 
  \item The I-GNN model \textbf{outperforms} the meta-learning-based methods under the inductive setting, particularly on datasets like Cora, CiteSeer and Corafull, while demonstrating competitive performance on other datasets. This can be attributed to the fact that meta-learning methods typically require a large number of samples to learn effectively.
\end{itemize}
\vspace{-30pt}
\vspace{-30pt}
\subsection{Further Analysis}

\begin{figure}[!htp]
\vspace{-10pt}
    \centering
    \includegraphics[width=12 cm]{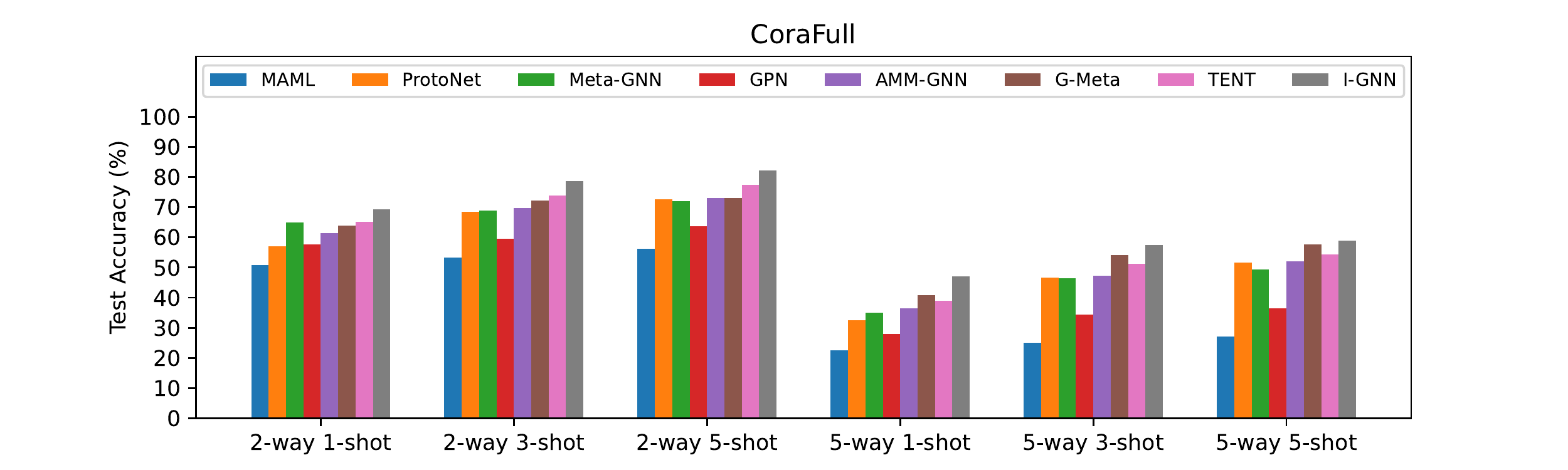}
    \caption{N-way K-shot results of CoraFull, Meta-Learning and I-GNN.}
    \label{fig2}
    \centering
    \includegraphics[width=6 cm]{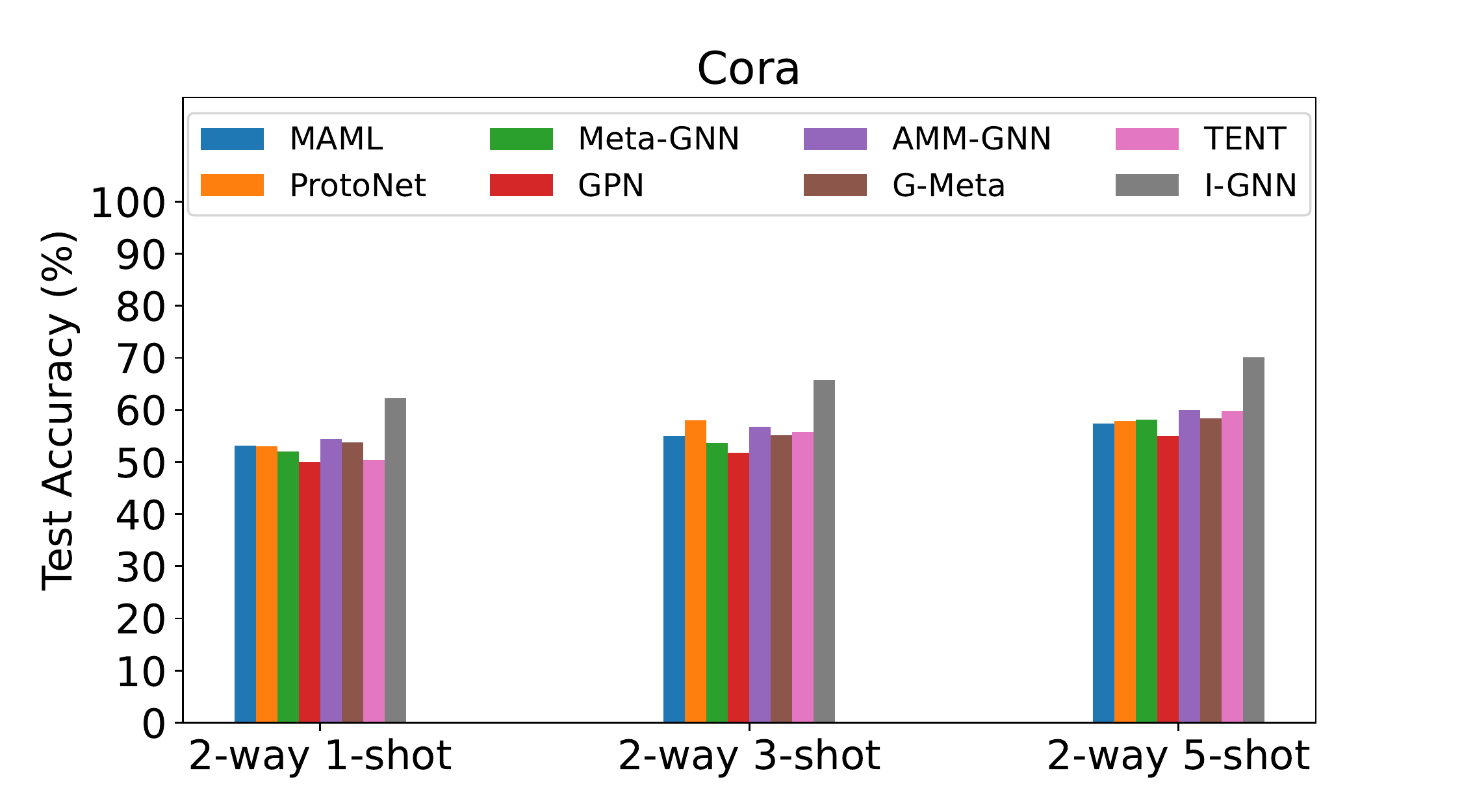}
    \includegraphics[width=6 cm]{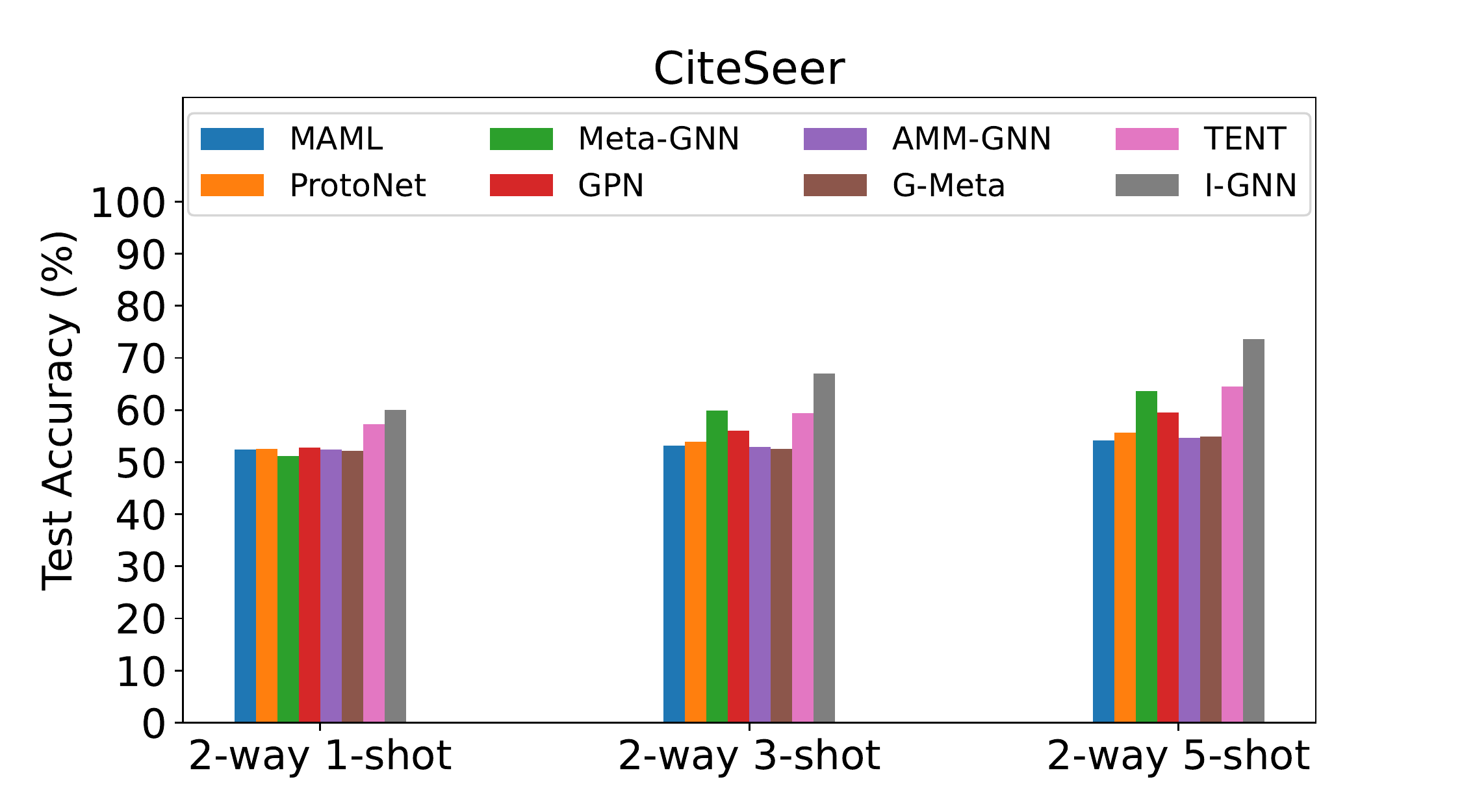}
    \caption{N-way K-shot results of Cora and CiteSeer, Meta-Learning and I-GNN.}
    \label{fig3}
    \vspace{-15pt}
\end{figure}

To make a direct comparison between the results of meta-learning methods and I-GNN, we present additional findings in Fig. ~\ref{fig2} and Fig. ~\ref{fig3}, which showcase the performance of all methods across different N-way K-shot settings. By analyzing these results, we can draw the following conclusions.
\begin{itemize}
  \item As N increases, the \textbf{performance of all methods deteriorates} due to the greater variety of classes within each meta-task. This increased complexity poses challenges for classification tasks, resulting in lower performance. Fig. ~\ref{fig2} demonstrates the impact of increasing N on the classification performance using the CoraFull dataset.
  \item The \textbf{performance improvement} of the I-GNN method compared to meta-learning methods on the Cora dataset, as shown in Fig. ~\ref{fig3}, is notable due to its smaller number of classes, allowing I-GNN to leverage structural information for better generalization. The meta-learning methods struggle to effectively utilize the available supervision information during training. 
\end{itemize}

\section{Conclusion}
In this paper, we investigate the performance of meta-learning methods in the inductive few-shot node classification tasks. While existing research primarily focused on the transductive setting, the inductive setting has received limited attention in the few-shot learning community. To bridge this gap, we conduct a comprehensive study of meta-learning for inductive few-shot node classification. Our empirical analysis reveals that most current meta-learning frameworks struggle in the inductive setting. To address this challenge, we propose a simple yet competitive baseline model called I-GNN. Experimental evaluations on five real-world datasets showcase the effectiveness of our proposed model. Our findings emphasize the need for further research in exploring the potential of meta-learning in the inductive setting, contributing to a more comprehensive understanding of few-shot node classification.

\end{document}